\documentclass[letterpaper, 10 pt, conference]{ieeeconf}  

\IEEEoverridecommandlockouts                              

\usepackage{graphicx}
\usepackage{color}
\usepackage{cite}
\usepackage{amsmath}
\usepackage{bm}
\usepackage{url}
\usepackage{float}
\usepackage[caption=false,font=footnotesize]{subfig}

\title{\LARGE \bf
Multimodal Uncertainty Reduction for Intention Recognition in Human-Robot Interaction
}

\author{Susanne Trick$^{1,2}$, Dorothea Koert$^{2,3}$, Jan Peters$^{2,3,4}$, Constantin A. Rothkopf$^{1,2,5}$
\thanks{\hspace*{-0.35cm} This work has been funded by the German Federal Ministry of Education}
\thanks{\hspace*{-0.35cm} and Research (BMBF) in the project KoBo34 (project no. 16SV7984).}
\thanks{\hspace*{-0.35cm} $^{1}$Psychology of Information Processing, TU Darmstadt, Germany}
\thanks{\hspace*{-0.35cm} $^{2}$Centre for Cognitive Science, TU Darmstadt, Germany}
\thanks{\hspace*{-0.35cm} $^{3}$Intelligent Autonomous Systems, TU Darmstadt, Germany}
\thanks{\hspace*{-0.35cm} $^{4}$MPI for Intelligent Systems, Tuebingen, Germany}
\thanks{\hspace*{-0.35cm} $^{5}$Frankfurt Institute for Advanced Studies, Goethe University, Germany}
\thanks{\hspace*{-0.35cm} {\tt\small susanne.trick@cogsci.tu-darmstadt.de}}%
}

\begin{document}

\maketitle
\thispagestyle{empty}
\pagestyle{empty}

\begin{abstract}
Assistive robots can potentially improve the quality of life and personal independence of elderly people by supporting everyday life activities. To guarantee a safe and intuitive interaction between human and robot, human intentions need to be recognized automatically. As humans communicate their intentions multimodally, the use of multiple modalities for intention recognition may not just increase the robustness against failure of individual modalities but especially reduce the uncertainty about the intention to be predicted. This is desirable as particularly in direct interaction between robots and potentially vulnerable humans a minimal uncertainty about the situation as well as knowledge about this actual uncertainty is necessary. Thus, in contrast to existing methods, in this work a new approach for multimodal intention recognition is introduced that focuses on uncertainty reduction through classifier fusion. For the four considered modalities speech, gestures, gaze directions and scene objects individual intention classifiers are trained, of which all output a probability distribution over all possible intentions. By combining these output distributions using the Bayesian method Independent Opinion Pool \cite{berger1985} the uncertainty about the intention to be predicted can be decreased.  The approach is evaluated in a collaborative human-robot interaction task with a 7-DoF robot arm. The results show that fused classifiers which combine multiple modalities outperform the respective individual base classifiers with respect to increased accuracy, robustness, and reduced uncertainty.

\end{abstract}

\vspace{0.2cm}
\section{Introduction}
A prevalent challenge for our society is an increasing number of elderly people in need of care facing a shortage of nursing staff \cite{iwreport}. A promising reaction to this is to investigate in technical solutions that can improve the quality of life of elderly people, not just by supporting caregivers but also by directly providing assistance to affected elderly people. Assistive robots are a potential answer to this. By facilitating harmful or arduous everyday life tasks they may enable even physically handicapped people to stay longer in their own habitual environments.\\
In order to guarantee trouble-free cooperation between human and robot, it is necessary that the robot automatically recognizes human intentions. Since humans make use of multiple modalities like speech, body language and situational clues for understanding intentions \cite{schrempf2007}, it is reasonable to also take advantage of multimodal data in automatic intention recognition.\\
Multimodal
\begin{figure}[t]
	\centering
	\includegraphics[width=\columnwidth]{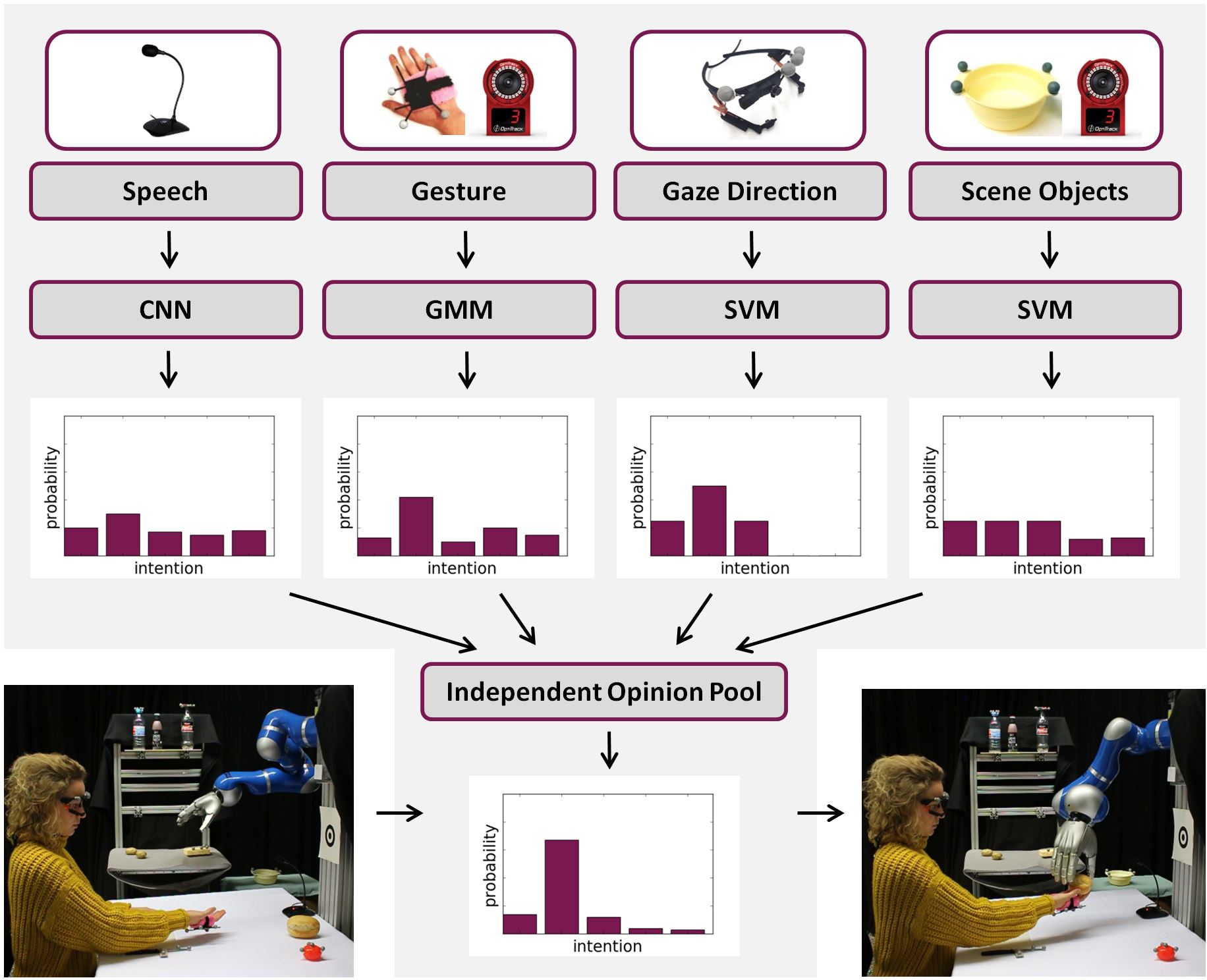}
	\caption{Assistive robots can potentially be applied to support elderly people in their daily lives, for which reliable intention recognition is necessary. In this work, we introduce a multimodal intention recognition system that fuses the modalities speech, gestures, gaze directions and scene objects. For all modalities individual classifiers are learned. Their output distributions are fused with the method Independent Opinion Pool in order to reduce the overall system's uncertainty.}
	\vspace{-0.5cm}
	\label{fig:introduction_overview}
\end{figure}
intention recognition is beneficial in two kinds of ways. On the one hand, it offers the possibility to compensate for limited or missing modalities, e.g. a speech disorder as a consequence of a stroke. On the other hand, through the integration of information of several modalities the uncertainty about the true intention to be predicted can be decreased \cite{ernst2002}. This is particularly important in the interaction between elderly and potentially vulnerable people and robotic systems to ensure safety. However, the reduction of uncertainty through the use of multiple modalities for intention recognition was not the focus of previous works.\\
\pubid{}
We propose a multimodal intention recognition system that focuses on uncertainty reduction (Fig \ref{fig:introduction_overview}).
Thereby, we consider the four modalities speech, gestures, gaze directions and scene objects as part of the situational context.
For all modalities an individual classifier is trained that returns a categorical probability distribution over all possible intentions. The resulting distributions are then fused with a Bayesian method for cue integration called Independent Opinion Pool \cite{berger1985}. Through the application of this method for classifier fusion the uncertainty about the intention to be predicted can be lowered even if the classifiers for the single modalities are individually inaccurate or uncertain. This is shown in a collaborative task where the human is supported by a 7-DoF robot arm in preparing some food in a kitchen scenario while human intentions are recognized online using the four considered modalities.\\
The rest of the paper is structured as follows. In Section \ref{sec:related_work} related work is discussed. Section \ref{sec:approach} explains the chosen approach for multimodal intention recognition including methods for classifier fusion and intention recognition from the four considered modalities. In Section \ref{sec:experiments} we show evaluations on multimodal data from a collaborative task with a 7-DoF robot and present corresponding results. Finally, we conclude with Section \ref{sec:conclusion} and discuss possible future work.

\vspace{0.2cm}
\section{Related Work} \label{sec:related_work}
Since the combined utilization of multiple modalities can improve the performance and robustness of a system \cite{li2010, nweke2018, rodomagoulakis2016}, there are several approaches that have already dealt with multimodal intention recognition in human-robot interaction.
The most popular combination of modalities in these works is the use of speech commands together with gestures \cite{mollaret2016, rodomagoulakis2016, stiefelhagen2004, vaufreydaz2016, zlatintsi2018}, sometimes additionally combined with head poses \cite{mollaret2016, stiefelhagen2004} or otherwise face information and movement speed of the respective person \cite{vaufreydaz2016}. Gaze was considered as a modality for intention recognition in two previous approaches, used either in combination with body pose \cite{yu2015} or with speech and buttons \cite{bannat2009}. Scene objects were only considered in relation to the human, not as passive parts of the scene \cite{dutta2016, dutta2018, kelley2012deep, kelley2012bayes}. Once they were additionally combined with body poses \cite{dutta2018}, once with the considered objects' states \cite{kelley2012deep, kelley2012bayes}. Two further proposed systems that considered completely different modalities are \cite{xu2015}, who combined force sensor and laser rangefinder data for recognizing motion intentions, and \cite{kulic2003}, who employed multiple physiological signals such as blood volume pressure for the automatic recognition of human approval of a robot's actions.
As can be seen, there are several systems that consider multiple modalities for intention recognition. However, none of the works has combined the four modalities speech, gestures, gaze directions and scene objects as it is done in this work.\\
Among all the works discussed so far, there are some that not only deal with intention recognition in human-robot interaction but especially address elderly assistance, which is of particular interest in this work.
Vaufreydaz et al. \cite{vaufreydaz2016} worked on the automatic detection of the intention to interact with a robot. Considered data were face size and position, speech, shoulder pose rotation and movement speed. These data were concatenated to form a single feature vector which was classified either by a Support Vector Machine or a Neural Network. Thus, feature fusion was conducted instead of classifier fusion which would fuse the outputs of individual classifiers.
Mollaret et al. \cite{mollaret2016} also dealt with the recognition of an intention for interaction with an assistive robot. Using head and shoulder orientation and voice activity the corresponding intention could be inferred with a Hidden Markov Model. Here, raw data instead of features were used for fusion but again not the outputs of individual classifiers.\\
Xu \cite{xu2015} proposed a walking-aid robot and in this context focused on recognizing human intentions in terms of intended walking velocities. Data were captured from force sensors and a laser rangefinder and the estimated velocities were fused with a Kalman filter. Even though here, outputs of individual estimators were fused, the study only dealt with a continuous size instead of category labels.\\
Rodomagoulakis et al.\cite{rodomagoulakis2016}, on the other hand, fused outputs of classifiers working on discrete categories. In order to enable people with limited mobility to interact with a robotic rollator, they recognized intentions considering speech and gestures. The respective classifiers' output scores for all possible intentions were fused by a weighted linear combination with tunable weights while the intention with the highest fused score was predicted.
Another work operating on classifier outputs for fusion by Zlatintsi et al. \cite{zlatintsi2018} proposed an intention recognition system for an assistive bathing robot based on speech and gestures. They applied a late fusion scheme meaning that an intention was chosen as the detected one if it was ranked highest by the speech classifier and among the two highest ranked intentions according to observed gestures. Although these two works fuse discrete classifiers' outputs, these outputs are not treated probabilistically, so uncertainty reduction is not possible.\\
In fact, all approaches regarded so far perform modality fusion for intention recognition. Some even do so by fusing outputs of individual classifiers. However, none of them considers uncertainty for fusion or attempts to reduce the uncertainty of the final decision. Instead, they are exclusively concerned with improving the system's accuracy and robustness.
This is also the case for works about multimodal intention recognition in other contexts than elderly assistance~\cite{yu2015}.\\
Some specialized approaches for audio-visual speech recognition \cite{gurban2008, liu2014} considered uncertainty by performing uncertainty-based weighting for the fusion of multiple classifiers' outputs. In these works, the respective two categorical probability distributions returned by two individual classifiers for audio and visual input were combined by a weighted sum. The respective weights were computed from the individual distributions' uncertainties, quantified e.g. with entropy \cite{gurban2008}. Consequently, the more uncertain distribution got the lower weight and by this had a lower influence on the fused distribution. Whereas it is desirable to consider uncertainty for determining the individual distributions' impact on the fusion result, a weighted sum of categorical distributions cannot reduce uncertainty, because it results in an average of the distributions which is less or equal certain per definition. However, uncertainty reduction in one of the biggest advantages of fusing different classifiers' distributions \cite{nweke2018}.\\
Consequently, another method is needed that not only determines each distribution's impact on the fused resulting distribution based on its uncertainty but also reduces the uncertainty of this fused distribution. A suitable method that meets this requirement is Independent Opinion Pool \cite{andriamahefa2017, berger1985} which we use in our approach (Sec \ref{sec:approach_iop}). It basically multiplies the individual probability distributions for fusion. The method has already been applied for different fusion tasks, among them the fusion of geological data from different measurement locations \cite{elsaesser2007}, of laser rangefinder data for semantic labeling of places \cite{shi2010} and of camera data for robust robot navigation \cite{stepan2005}. However, Independent Opinion Pool has not been used for multimodal intention recognition so far in order to explicitly reduce uncertainty.\\
All in all, to the best of our knowledge there is no work that uses the four modalities we consider for intention recognition together with a method that targets uncertainty reduction.

\vspace{0.2cm}
\section{Multimodal Intention Recognition} \label{sec:approach}

In this work, an approach for multimodal intention recognition is introduced which focuses on reducing the uncertainty about the intention to be predicted. Since we represent intentions as discrete categories, recognizing them can be seen as a classification problem. Our approach applies classifier fusion which fuses the outputs of individual and independent base classifiers instead of e.g. fusing directly the raw data or respective feature vectors. For this reason, for each of the four considered modalities, a classifier was trained on its own data from the respective modality. Each individual classifier returns a categorical probability distribution over all possible intentions as output, which contains a probability for each possible intention. All of these base classifiers could perform intention recognition on their own. However, their output distributions are fused in order to decrease uncertainty and improve performance. An overview of the proposed approach is shown in Figure 1.

\subsection{Classifier Fusion with Independent Opinion Pool} \label{sec:approach_iop}

\begin{figure}
	\centering
	\includegraphics[width = \columnwidth]{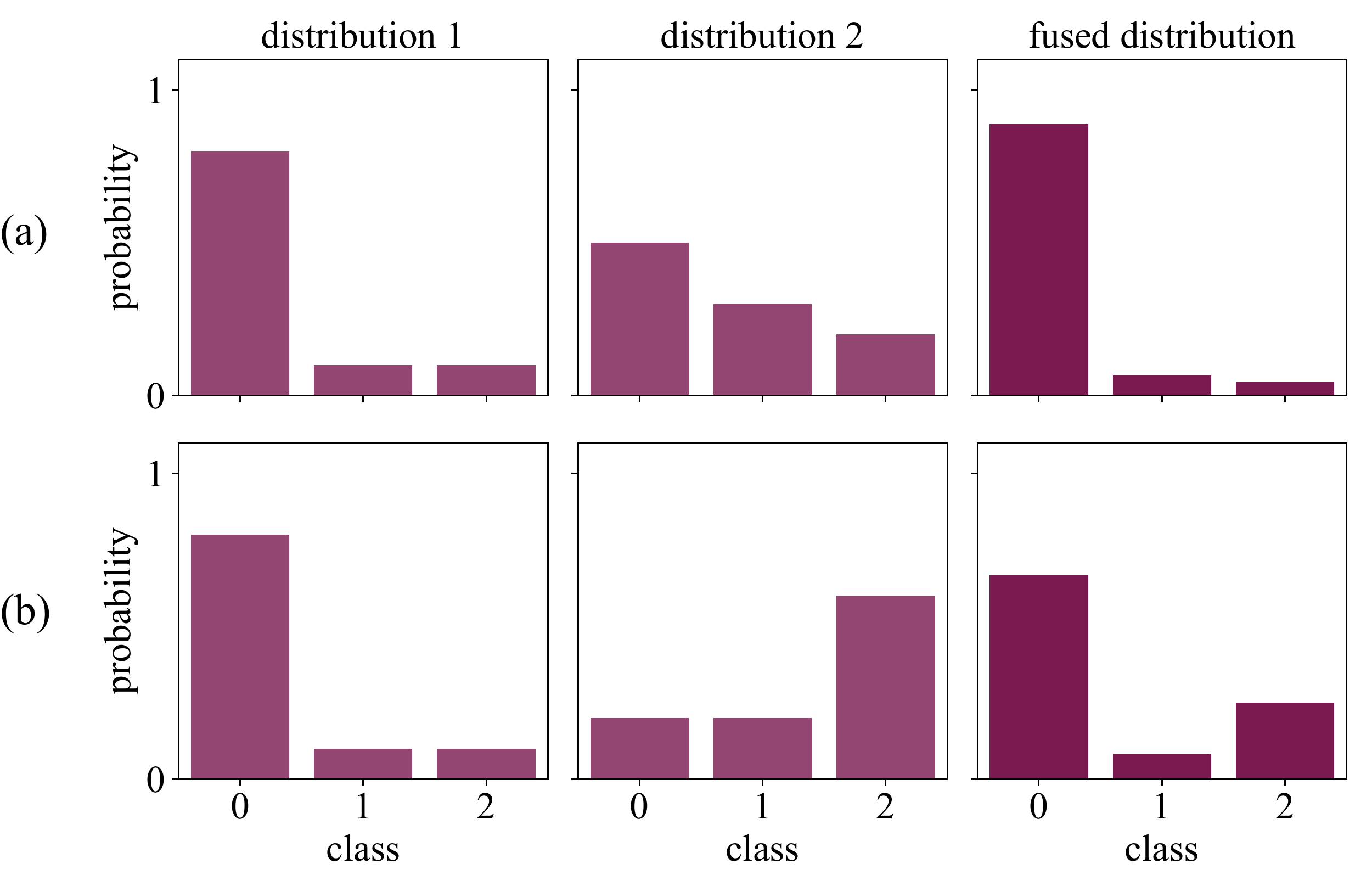}
	\vspace{-0.6cm}
	\caption{(a) When used for classifier fusion, Independent Opinion Pool leads to reinforcement, which means that two non-conflicting distributions result in a more certain fused distribution. (b) It also leads to mitigation, meaning that the fusion of two conflicting distributions causes an increased uncertainty. Meanwhile, the more certain distribution 1 has a higher impact on the fused distribution and is thus decisive.}
	\vspace{-0.3cm}
	\label{fig:iop_examples}
\end{figure}

Our principal motivation for combining multiple modalities is uncertainty reduction. First, the fusion of two non-conflicting distributions, in the most extreme case two equal distributions, should result in a fused distribution with a lower entropy than those of the respective base distributions. Second, the uncertainty of each base distribution, e.g. in terms of its entropy, should determine the influence of the distribution on the fused result in a way that an uncertain distribution's influence is lower.\\
In order to achieve uncertainty reduction as it is described above, we apply Independent Opinion Pool \cite{berger1985} for fusion of the $n$ categorical probability distributions $P(y|x_i)$ over intentions $y$ given modality data $x_i$. This method assumes conditional independence of the base classifiers given the true class label which is the true underlying intention in our case. Furthermore a uniform distribution over all possible classes $P(y)$ is assumed a priori. By applying Bayes rule with these assumptions, the fusion can be conducted by simply multiplying the underlying base distributions returned by each classifier and renormalizing the resulting categorical distribution so that it sums to one,
\begin{align}
P(y|x_1, ..., x_n) \propto \prod_{i=1}^{n}{P(y|x_i)}.
\end{align}
Two advantages of this method for fusion are reinforcement and mitigation \cite{andriamahefa2017}. Reinforcement describes the first criterion for uncertainty reduction we set in the previous paragraph. In case the distributions returned by the individual base classifiers are non-conflicting and thus predict the same class, the uncertainty and with it the entropy of the resulting fused distribution is reduced compared to those of the base distributions (Fig \ref{fig:iop_examples}a). Mitigation means that conflicting base distributions cause a fused distribution with a higher uncertainty (Fig \ref{fig:iop_examples}b). This might seem to be a contradiction to this work's goal of uncertainty reduction but is indeed desirable as in case that cues from different modalities are conflicting the resulting fused distribution should reflect this.\\
The second criterion for uncertainty reduction is also accomplished. Using Independent Opinion Pool for fusion, each base distribution's uncertainty determines its impact on the fusion result. The fusion impact of an uncertain base distribution is lower than that of a more certain one (Fig \ref{fig:iop_examples}b). Hereby, the fusion impact of each base distribution is only dependent on its current uncertainty and is thus recomputed online for every new multimodal data example.
\vspace{0.1cm}

\subsection{Classifiers for Single Modalities}

The individual classification methods we use for intention recognition from the four modalities speech, gestures, gaze directions and scene objects are presented in the following.
However, it is not the focus of this work to develop new methods for classifying intentions from the different single modalities' data. Instead, the power of multimodality for uncertainty reduction is demonstrated. Thus, the base classifiers for the considered modalities are designed as simple as possible and build upon existing methods. In fact, they can be easily replaced by any other classifiers which output categorical distributions, and additional classifiers might be added sequentially in order to further improve the system.

\subsubsection{Speech}

Speech is a meaningful modality for intention recognition as it is effortless and intuitive for humans \cite{liu2016}. For intention recognition from speech, keyword spotting is applied instead of continuous speech recognition. This enables the recognition of simple keywords that are related to intentions while simultaneously allowing humans to flexibly formulate the command sentences. Since a probability distribution over all keywords is required as output, which in many popular frameworks is not available, an open-source framework called Honk \cite{tang2017} is used. Honk builds upon a Convolutional Neural Network proposed by Sainath and Parada \cite{sainath2015} that consists of two convolutional layers and one final softmax layer. As input for the network, Mel-Frequency Cepstrum Coefficient features are used. Its implementation is realized with PyTorch.\\
For training, we recorded keyword utterances from 16 people, eight of them female and eight male, while each keyword, e.g. "bowl" for the intention to get a bowl, was repeated ten times. In addition to recordings of the keywords of interest also eleven other words were recorded that are likely to be part of possible command sentences in order to reduce false alarms. 15\% of the training data were taken from these unknown words, another 15\% were taken from example recordings for silence, e.g. noise sounds. To increase robustness, these noise sounds were also added to the training examples with a probability of 0.8. 80\% of all data were taken for training and 10\% each for testing and validation.\\
Since the network is trained on single keywords but keywords need to be detected within complete command sentences, multiple probability distributions are obtained for one query sentence. To combine them, for each intention the maximal probability value in all distributions is taken to constitute the final probability distribution. This is motivated by the assumption that each sentence contains only one keyword to which the highest probability should be assigned.\\
The device for recording speech is a USB microphone (Klim) that captures mono sound with a sample rate of 16000 Hz. The recorded data are saved in 16-bit little-endian PCM-encoded wave format.

\subsubsection{Gestures} \label{sec:approach_gestures}
Since a majority of human communication is nonverbal, gestures provide valuable information about intentions, especially when referring to objects \cite{canal2015}. Here, we realize intention recognition from gestures based on the method Mixture of Interaction Primitives \cite{ewerton2015}. 
In this method, gesture trajectories are represented with Probabilistic Movement Primitives (ProMPs) \cite{paraschos2013} which approximate each trajectory position by a linear combination of Gaussian basis functions and weights $\bm{w}$. Using this representation one can learn a probability distribution over multiple demonstrated trajectories. Mixture of Interaction Primitives \cite{ewerton2015} extends ProMPs to be usable with multiple gestures and two interacting agents, e.g. a human and a robot. For this, a Gaussian Mixture Model (GMM) over human and robot trajectories represented as ProMPs is learned, in which each mixture component represents one interaction pattern between human and robot. In addition to inferring a learned gesture from an observed human trajectory, the method is also able to estimate the most likely response trajectory of the robot conditioned on an observed human trajectory. By this, the robot's movement can be adapted to the actually shown human gesture, e.g. with respect to a common end position in a handover task. For more details on the used approach the interested reader is referred to \cite{ewerton2015}.\\
We apply Mixture of Interaction Primitives for intention recognition by representing gestures together with corresponding robot reaction trajectories as ProMPs and learning a respective GMM with one mixture component for every intention. Thus, in addition to recognizing intentions from human gestures, we can also generate a corresponding robot trajectory as a reaction to the intention recognized by the fused classifier considering all four modalities.\\
The parameters of the GMM are originally learned from unlabeled data with the Expectation Maximization algorithm \cite{ewerton2015}. However, as we work with labeled data, we estimate the GMM's parameters with Maximum Likelihood Estimation. In addition, just the last point of the observed trajectory is taken for gesture classification which is sufficient for differentiating the gestures in our experiment.\\
For training of the gesture classifier, 30 examples of reaching motions were demonstrated by one human subject. The resulting trajectories were captured with the motion tracking system Optitrack which uses cameras and passive reflective markers attached to the human wrist. For training of the robot reaction movements, kinesthetic teaching was applied.

\subsubsection{Gaze Directions}
Previous studies revealed fixations to be strongly task-dependent and predictive for future actions and intentions \cite{admoni2016, huang2015, mennie2007, rothkopf2007}, which motivates us to infer human intentions from gaze directions. For this, we apply a Support Vector Machine (SVM) that is inspired by two existing approaches about intention recognition from gaze \cite{huang2015, admoni2016}. Considered features are the distances between the human's 3D gaze vector and all locations of interest in the scene, which are mainly object locations. The mean gaze vector is computed from the last 900 samples of the recorded gaze directions during a trial. The motivation for working with this mean here is that it indirectly includes information about the most recently fixated location and the number and duration of fixations towards this location, which were all stated to be important features for intention recognition in \cite{huang2015}. Additionally considering the distances between locations of interest and the gaze vector is an idea presented in \cite{admoni2016}. While using the described features, a multiclass SVM with linear kernel was trained using the package sklearn. In contrast to just the predicted class label, which is the usual output of an SVM, this package also provides a probability distribution over all possible labels as output.\\
The used training data included 30 labeled examples per intention, each with a duration of five seconds and consisting of 1250 samples. For recording, one person was seated in the scenario setup described in Section \ref{sec:experiments} and asked for several robot assists, e.g. a handover of an object, while shifting its gaze towards the location of interest, i.e the object itself. Gaze direction is recorded monocular by a head-mounted eye tracker (Pupil Labs) which uses infrared lights and eye cameras for inferring the pupils' positions and computing the gaze vector.
The device is additionally equipped with reflective markers in order to be trackable by the Optitrack system also used for gesture tracking, because we need the gaze vector in scene coordinates rather than just related to the eye tracker itself. For integration of the eye tracker in the overall system an open-source ROS plugin is used \cite{pupilrosplugin}.

\subsubsection{Scene Objects}
Scene objects are objects that are passive parts of the scene but can still give hints about what the intention of the human could be \cite{bach2014}. For estimating intentions from such objects we build upon an approach that deals with scene classification from observed objects \cite{luo2016}. The reason for this choice is that other approaches which directly deal with intention recognition only consider objects that are manipulated by a human \cite{dutta2016, dutta2018, kelley2012deep} rather than passive scene objects. Thus, an SVM was chosen as classifier type with input feature vectors containing the horizontal distances of all available scene objects to a pre-defined center point on a working area in front of the human, which is basically the front part of a table. Objects that are positioned outside this working area are set to the same pre-defined value. Other possible features, such as the raw positions of objects on the working area or just Boolean values indicating whether an object is in the working area or not, performed worse than the chosen approach. The multiclass SVM is again implemented using the package sklearn and again a probability distribution over all intentions is returned instead of just one predicted intention.\\
Training was conducted on 50 recordings of different scene object placements for each intention, e.g. a glass for the intention to get some coke. Object positions are gathered with the camera-based motion tracking system Optitrack. For this, each scene object is equipped with four markers in a unique geometric pattern that enables the system to distinguish between objects and record their position. Used scene objects in our human-robot interaction scenario are a cutting board, a tomato, a bowl, a bottle of each coke and water, a sponge, a glass and a knife.\\

\section{Experimental Evaluation} \label{sec:experiments}

\begin{figure}[t]
	\centering
	\includegraphics[width=\columnwidth]{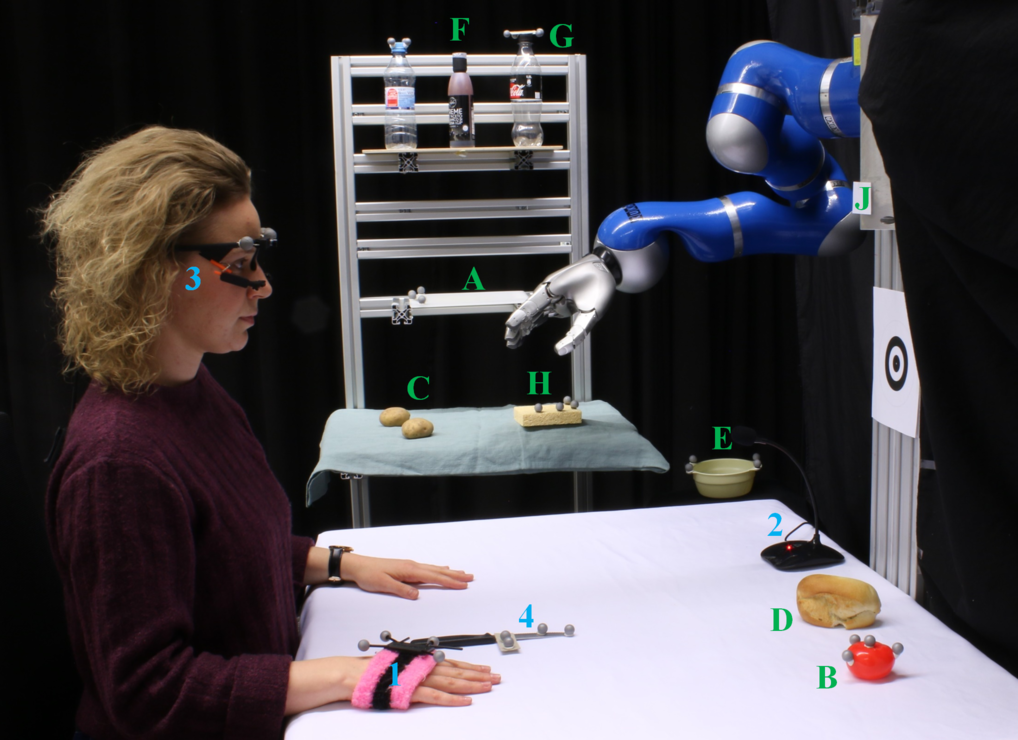}
	\caption{The kitchen scenario the fusion system was evaluated in. The human can request different actions from the robot by displaying intentions through the four considered modalities. For capturing gestures and scene objects the human's hand (1) and the scene objects (e.g. 4) are equipped with markers. Speech is captured with a microphone (2) and gaze with a head-mounted eye tracker (3). The ten recognizable intentions are the handover of the board (A), tomato (B), potato (C), roll (D), bowl (E), dressing (F), coke (G) or towel (H). Additionally, there is the intention to stand up for which location (J) is fixated.}
	\vspace{-0.6cm}
	\label{fig:scenario_setup}
\end{figure}
\begin{figure}[t]
	\centering
	\includegraphics[width=\columnwidth]{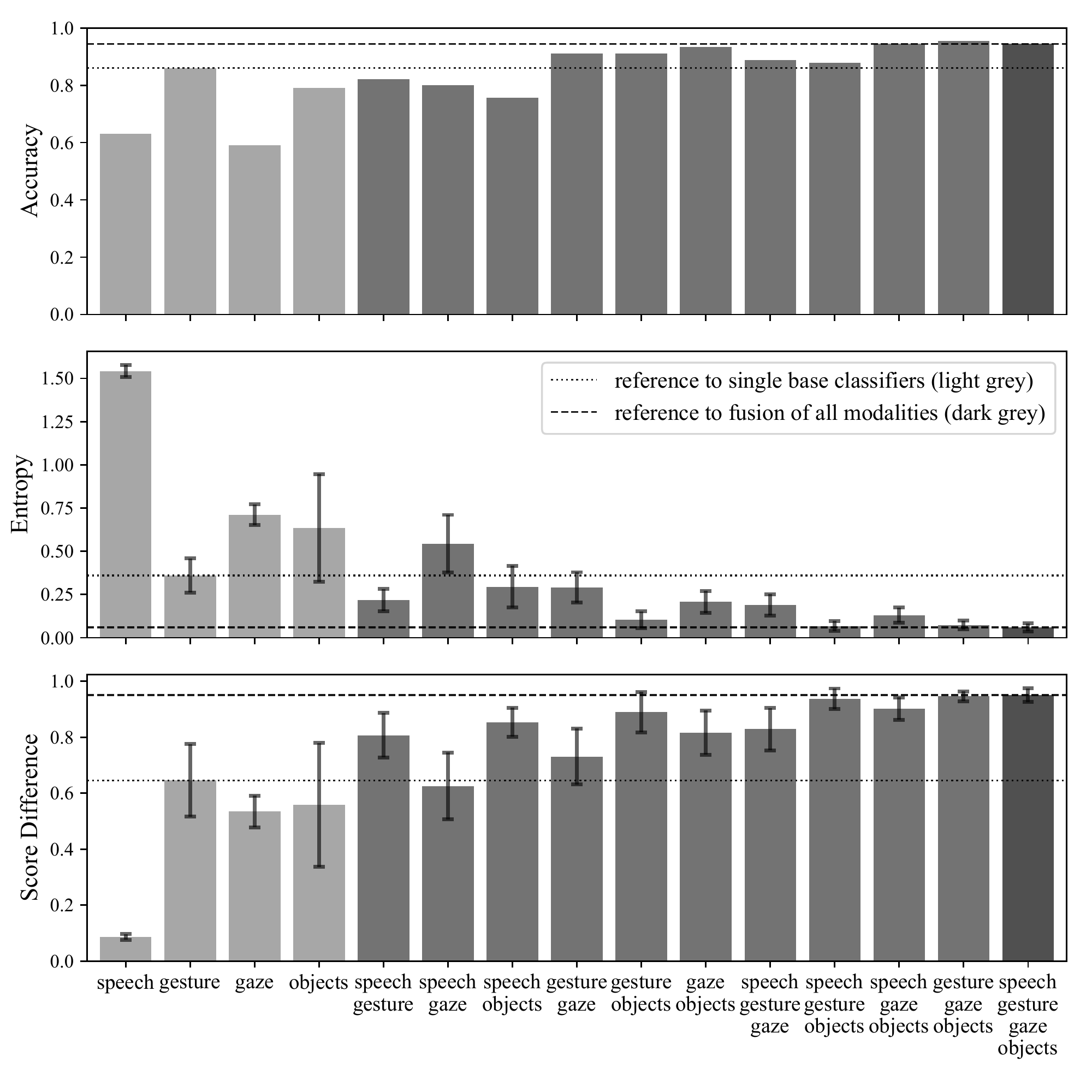}
	\vspace{-0.7cm}
	\caption{Comparison of all possible combinations of base classifiers regarding accuracy, mean entropy and score difference. Corresponding variances are plotted as error bars. It is seen that uncertainty is reduced through classifier fusion, in particular it is lowest for the fusion of all four modalities, according to both measures entropy and score difference. Accuracy is also increased through fusion compared to the single classifiers.}
	\vspace{-0.4cm}
	\label{fig:acc_ent_diff}
\end{figure}
\begin{figure*}[h]
	\centering
	\includegraphics[width=\textwidth]{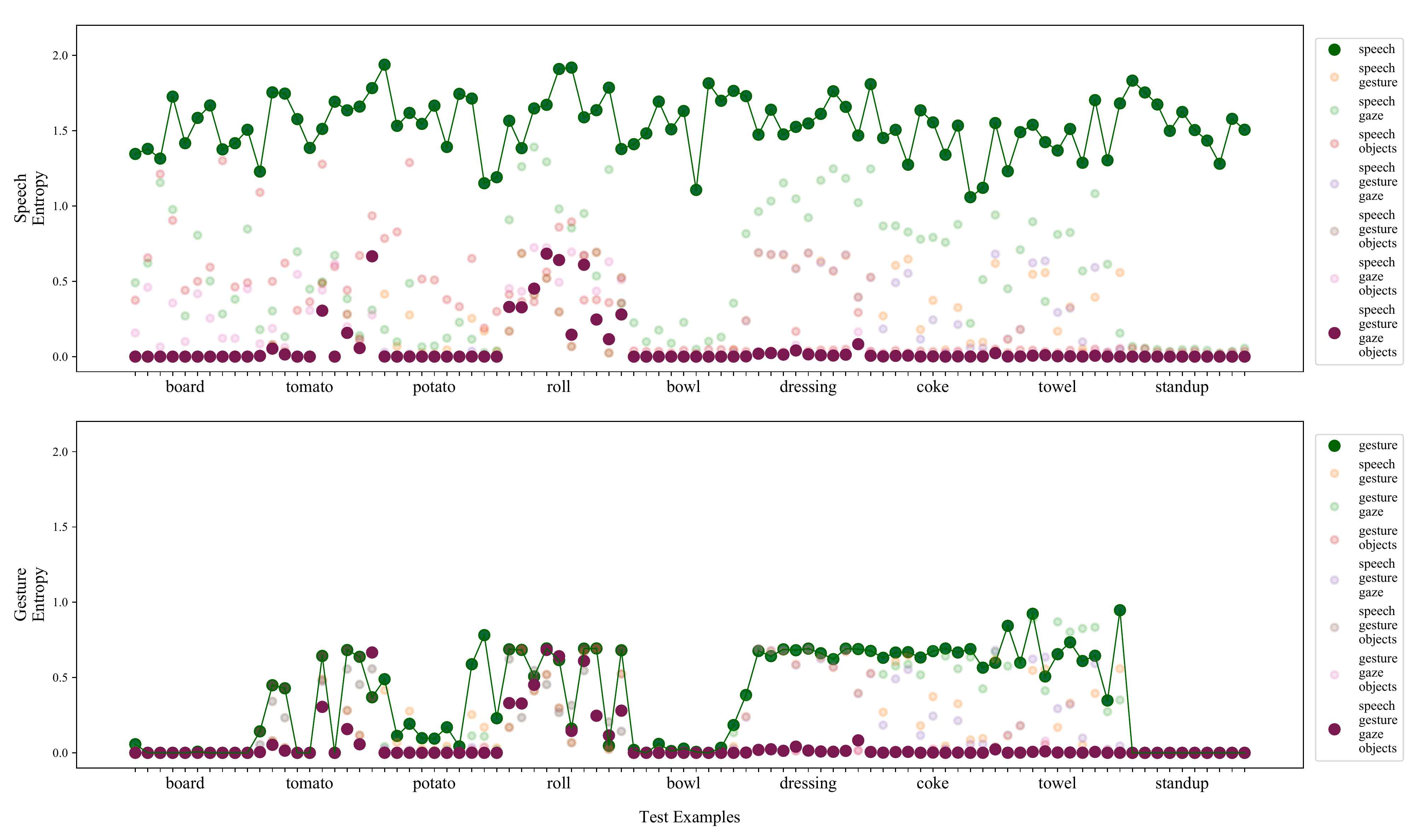}
	\vspace{-0.6cm}
	\caption{Entropies of all distributions resulting from classifier combinations including the speech (top) or gesture (bottom) classifier over all 90 test examples for the nine intentions. We see that for a large majority of examples already combinations of two or three classifiers reduce the uncertainty compared to the uncertainty of the single speech or gesture classifier. The fusion of all four modalities causes the strongest uncertainty reduction. The intentions tomato and roll show higher uncertainties since the base classifiers often confound them.}
	\label{fig:scatter_entropy}
\end{figure*}
\begin{figure*}[h]
	\centering
	{\subfloat[]{\includegraphics[width=0.16\textwidth]{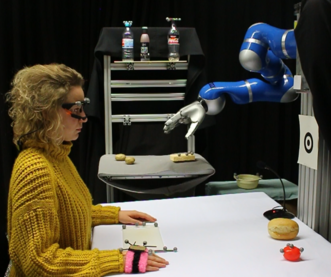}}}
	{\subfloat[]{\includegraphics[width=0.16\textwidth]{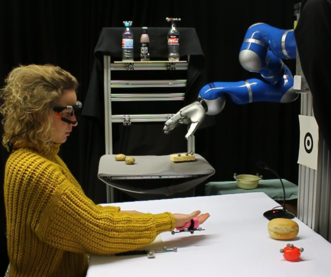}}}
	{\subfloat[]{\includegraphics[width=0.16\textwidth]{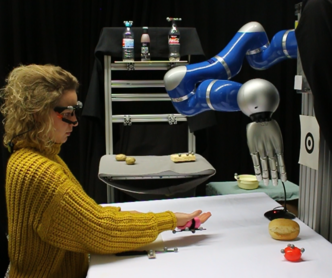}}}
	{\subfloat[]{\includegraphics[width=0.16\textwidth]{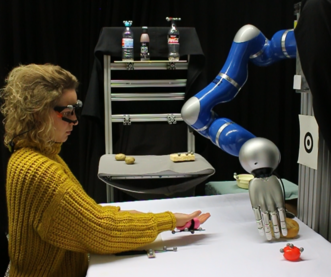}}}
	{\subfloat[]{\includegraphics[width=0.16\textwidth]{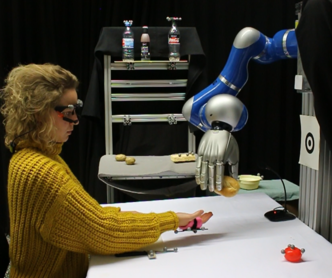}}}
	{\subfloat[]{\includegraphics[width=0.16\textwidth]{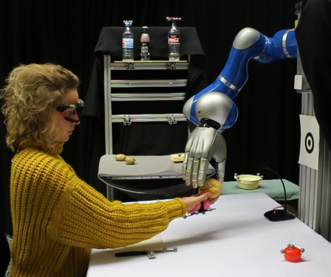}}}
	\caption{The human shows the intention roll by uttering a command containing the word "roll", reaching out its arm for the roll, fixating the roll and having placed the scene objects board and knife in the working area (b). As soon as this intention is recognized, the robot moves towards the roll (c), grasps it (d) and subsequently executes the inferred trajectory (e) to hand it over to the human (f).}
	\label{fig:roll_workflow}
	\vspace{0cm} 
\end{figure*}
For evaluation of the proposed multimodal intention recognition system we chose a kitchen scenario in which a 7-DoF robot arm assists a human in preparing some food, e.g. a salad (Fig \ref{fig:scenario_setup}). The human sits at a table with several task-relevant objects placed around him that are not easily reachable from a seated position. The robot can assist by handing over requested objects or helping to stand up by reaching out his arm as a prop. The nine recognizable intentions are to receive a cutting board, a tomato, a potato, a roll, a bowl, a bottle of dressing or coke or a towel and to get support for standing up. These intentions are deliberately chosen to be ambiguous, e.g. with respect to their positioning or sound, in order to obtain uncertain results when using the individual base classifiers only.
For all intentions ten example data sets were recorded, each consisting of a speech recording, the last point of the shown hand gesture, a list of the shown gaze directions and the positions of the scene objects of interest. Consequently, 90 multimodal examples from one human subject are available for evaluating the proposed intention recognition system.\\
Two measures are chosen to quantify the uncertainty of a categorical distribution. First, Shannon entropy is used, which is a well-known measure for the uncertainty of a distribution. It is maximal for a maximally uncertain distribution, that is uniform over all intentions, and minimal for a maximally certain distribution that assigns all probability mass to one intention.
Second, a measure called score difference \cite{potamianos2000} is applied for measuring uncertainty. It computes the difference between the highest and the second highest probability in the distribution. Thus, in contrast to entropy, score difference does not consider the complete distribution's uncertainty but quantifies the actual uncertainty in making a hard decision for one intention. Thereby, a low score difference indicates a high uncertainty.\\
Although uncertainty reduction is the focus of our approach, it is also essential to guarantee that the intentions are classified correctly. Thus, in addition to entropy and score difference also the accuracy of the multimodal and unimodal classifiers is evaluated.
Figure \ref{fig:acc_ent_diff} shows accuracy, entropy and score difference of the underlying fused distributions for all possible combinations of base classifiers as well as for the four single base classifiers. As can be seen, the accuracy of the fused result combining all modalities (0.94) is higher than that of the single classifiers (speech: 0.63, gesture: 0.86, gaze: 0.59, objects: 0.79). Only one other combination of base classifiers, namely gesture-gaze-objects, has a slightly higher accuracy. In general, eight out of eleven combinations of base classifiers result in a higher accuracy than all of the base classifiers do. This already indicates the superiority of multimodal over unimodal intention recognition.\\
When additionally considering uncertainty, both measures entropy and score difference show the lowest mean uncertainty for the fusion of all four modalities compared to all other possible combinations. Two other combinations of base classifiers, namely speech-gesture-objects and gesture-gaze-objects, show a similarly low uncertainty which yet is still higher than that resulting from fusing four modalities. In general, it can be seen that except from one classifier combination the uncertainties of the fused distributions are considerably lower than that of all single base classifiers. The only exception is the fusion of speech and gaze classifier which is slightly more uncertain than the most certain base classifier, the gesture classifier. Yet, its uncertainty is reduced in comparison to the two actually fused individual classifiers for speech and gaze. And it needs to be taken into account that these two classifiers are the least accurate and most uncertain of all four classifiers. This demonstrates the power of multimodal classifier fusion for intention recognition as proposed here. Even inaccurate and uncertain classifiers like speech and gaze classifier contribute to uncertainty reduction and better performance when added to a multimodal intention recognition system. Moreover, also combinations of base classifiers including less than all four modalities already improve performance and reduce uncertainty.\\
So far, the results convey the impression that the proposed multimodal approach for intention recognition reduces the uncertainty of the overall system. However, just means and variances of entropy and score difference over all test examples were taken into consideration. We additionally need to analyze whether uncertainty reduction is also accomplished for individual fusion examples. For speech and gesture classifier, Figure \ref{fig:scatter_entropy} shows the uncertainties in terms of entropy of the generated categorical distributions of all 90 test examples, differentiated according to whether they are generated by just the single classifier or by a fusion of multiple classifiers.\\
One can see that for the most uncertain classifier, the speech classifier, for all recorded test examples the fused distributions are always less uncertain than the single base distribution is, no matter how many of the three other modalities are added for fusion. In particular, the entropy of the distribution fused from all four modalities is lowest for nearly all examples. The only exceptions are examples from the intentions tomato and roll. Yet, this is easily explainable as these two intentions are often confounded by the base classifiers, which leads to conflicting base distributions. This in turn results in a higher uncertainty of the fused distribution which is desirable as different opinions of base classifiers should be reflected in the resulting fused distribution.\\
For the most certain one of the base classifiers, the gesture classifier, similar results can be shown, however not as strong as for the speech classifier. The gesture classifier is already quite certain on its own which is seen on much smaller overall entropy values. Apart from some exceptions, again, the fused distributions from all possible combinations with the other three modalities are less uncertain than the single gesture classifier, and in a majority of cases the fusion of all four modalities results in the lowest entropies near to zero. In contrast to the speech classifier seen before, for the gesture classifier there are some examples with higher entropies for the distribution resulting from the fusion of four modalities compared to the single classifier's distribution, but all these examples repeatedly come from the two ambiguous intentions tomato and roll.\\
These cases are especially interesting as the examples which are classified incorrectly by the fused distribution that combines all possible modalities mostly are examples for the intention roll. Consequently, the intention with the most uncertain fused distributions is also the intention with the most incorrect classifications. As an uncertain misclassification is more desirable than a certain one, this is desirable behavior.\\
Our proposed intention recognition system was not just evaluated quantitatively on recorded multimodal data but also online in a real interaction with the 7-DoF robot arm. For this, the kitchen task was performed cooperatively by a human and a robot. This means that in order to prepare a salad the human expresses the different intentions using the four modalities and after having recognized the correct intention the robot reacts accordingly. As a reaction, the robot moves to the respective location for this intention, e.g. the position of a requested object, and grasps it. Subsequently, it executes a trajectory in order to hand over the respective object or help the human to stand up. This trajectory was learned from demonstrations and is conditioned on the last point of the shown human movement as was explained in Section \ref{sec:approach_gestures}. This means that the most likely robot movement given the last point of the observed human trajectory is executed, which leads to an adaption of the robot movement to the human. This is especially beneficial for our handover tasks. The complete interaction process is shown exemplarily in Figure \ref{fig:roll_workflow} for the intention to get a roll. The human expresses this intention by uttering a command containing the word "roll", reaching out its arm in the roll's direction, fixating it and having placed the scene objects board and knife in the working area. The robot recognizes the correct intention and subsequently moves towards the roll, grasps it and hands it over to the human by executing the inferred trajectory.\\
As well as on recorded data, also in online interaction with a real robot it could be shown for all considered intentions that the proposed multimodal intention recognition system using speech, gestures, gaze directions and scene objects works.
\section{Conclusions} \label{sec:conclusion}
In this work, we introduce a multimodal approach for intention recognition to be applicable in elderly assistance. In contrast to existing works, we focus on uncertainty reduction in a way that the combination of modalities makes the system more certain about the intention to be predicted. For this, the categorical output distributions of individual classifiers for the four different modalities speech, gestures, gaze directions and scene objects are fused. We evaluate our approach in a cooperative kitchen task between a human and a 7-DoF robot arm. The results show that uncertainty can be decreased through the use of multiple modalities. Even very inaccurate and uncertain classifiers can contribute to uncertainty reduction, better performance and robustness when added to a multimodal system.\\
This shows that the proposed approach allows well-performing and certain intention recognition using simple and easily trained base classifiers that only require a low amount of training data. Additionally, it is particularly important for elderly assistance since even if complex classifiers are available they might be challenged by data from elderly people, which can increase their uncertainty and error rate.\\
In addition to utilizing uncertainty information for an optimal fusion of multiple modalities, for future work we plan to exploit the knowledge of the decision uncertainty in a way that the robot reacts according to its uncertainty about the situation. Another interesting line for future work we currently investigate is the inclusion of a reliability measure into the fusion mechanism. Because, although it is desirable that for every new decision the uncertainty of all base classifiers determines their impact on the fused result online, some additional knowledge about the individual performance of the classifiers could further improve the system.

\addtolength{\textheight}{-12cm}   


\bibliography{references}{}
\bibliographystyle{IEEEtran}

\end{document}